\pdfoutput=1
\documentclass[11pt]{article}
\usepackage[final]{acl}
\usepackage{times}
\usepackage{latexsym}
\usepackage{soul}
\usepackage[T1]{fontenc}
\usepackage[utf8]{inputenc}
\usepackage{microtype}
\usepackage{inconsolata}
\usepackage{graphicx}
\usepackage{enumitem}
\usepackage{amsfonts}
\usepackage{adjustbox}
\usepackage{booktabs}
\usepackage{multirow}
\usepackage{bm}
\usepackage{diagbox}
\usepackage{subcaption}
\usepackage{algpseudocode}
\usepackage{algorithm}
\usepackage{listings}
\usepackage{multicol}
\usepackage{caption}
\usepackage{amsmath}
\usepackage{amssymb}

\DeclareMathOperator*{\argmin}{arg\,min}

\title{A Training Data Recipe to Accelerate\\A* Search with Large Language Models}

\author{Devaansh Gupta\\
  \texttt{devaansh@cs.ucla.edu} \\
  Nanyang Technological University \\
  University of California, Los Angeles \And
  Boyang Li \\
  \texttt{boyang.li@ntu.edu.sg} \\
  Nanyang Technological University
  }

\begin{document}
\maketitle
\begin{abstract}
Combining Large Language Models (LLMs) with heuristic search algorithms like A* holds the promise of enhanced LLM reasoning and scalable inference. To accelerate training and reduce computational demands, we investigate the coreset selection problem for the training data of LLM heuristic learning. Few methods to learn the heuristic functions consider the interaction between the search algorithm and the machine learning model. In this work, we empirically disentangle the requirements of A* search algorithm from the requirements of the LLM to generalise on this task. Surprisingly, we find an overlap between their requirements; A* requires more accurate predictions on search nodes near the goal, and LLMs need the same set of nodes for effective generalisation. With these insights, we derive a data-selection distribution for learning LLM-based heuristics. On three classical planning domains, maze navigation, Sokoban and sliding tile puzzles, our technique reduces the number of iterations required to find the solutions by up to $15\times$, with a wall-clock speed-up of search up to $5\times$. The codebase is at \url{https://github.com/devaansh100/a_star}.
\end{abstract}

\section{Introduction}
Contrary to the view of Large Language Models (LLMs) as a monolithic paradigm for intelligence, the dual-process theory of cognitive science \cite{Stanovich_West_2000,thinking-fast-and-slow:2011} posits that human cognition consists of two closely collaborating systems, System 1 and System 2. System 1 exhibits typical traits of statistical learning such as fast inference and slow adaptation to novel problems. In comparison, System 2 can solve novel problems and excels at logical reasoning, but its inference speed is slow. 

Recent analyses analogize LLMs to System~1~\cite{saha2024system1xlearningbalancefast,wang2024litesearchefficacioustreesearch}, as LLMs perform poorly at novel, out-of-distribution problem formulations \cite{wu2024reasoning} or problems that require planning and reasoning \cite{valmeekam2023planning,tiong-etal-2024-measuring,cheng2024inductivedeductiverethinkingfundamental,kambhampati2024llmscantplanhelp}. On the other hand, tree-search methods like A* \cite{a-star-1968} and variants (e.g., \citealt{IDA-1985,MCTS-2006}), provide classic solutions to logical reasoning and planning, but they are unable to learn from past experiences and limited in speed due to sequential dependencies. Though it has been speculated that Artificial General Intelligence requires both System 1 and System 2 capabilities \cite{saha2024system1xlearningbalancefast,yu2024distilling21}, how to fruitfully combining LLMs with search techniques remains an open problem.

We study the problem of using LLMs to learn A* heuristics, which are functions that estimate the distance from a search node to the goal state. However, it can be computational demanding to train LLMs and to generate training data, as ground-truth labels for training can only be obtained from successfully solved problems.  

With this paper, we aim to improve the efficiency of heuristic learning by selecting a small subset of training data, known as the coreset, which would lead to near-identical A* performance as the whole dataset. To the best of our knowledge, no previous work investigated the coreset selection problem for A* heuristic learning. 

A complication of coreset selection in the A* + LLM setup is that the two algorithms may impose different requirements on training data. In this work, we attempt to disentangle and individually quantify the requirements of the two algorithms. We empirically test how different training data would change the generalisation of the LLM, and how A* reacts to generalisation errors in different positions of the search trajectory. 

We divide the training trajectory into three equally sized portions: the beginning, the middle, and the end. First, we evaluate their effectiveness as LLM training data. This is inspired by research using training data difficulty as a metric for coreset selection \cite{paul2021deep}. 
A natural definition for difficulty in A* is the distance to goal, which indicates how many decisions must be made before reaching the goal. Intuitively, it should be more difficult to guess the exact distance to goal at a given search node if the search node is in fact farther away from the goal. Further, to simulate the effect of LLM noise on A*, we inject random errors into oracle heuristic values in the three portions and observe effects on the search length. 

We obtain interesting and unexpected findings. For the LLM, training on the last portion, where the search node is closest to the goal and the distances are easiest to fit, leads to the best generalisation among the three portions. Unexpectedly, A* demonstrates a similar behavior; correct predictions on the end portion are the most beneficial to search efficiency, even though one might expect earlier decisions to be more important in pruning search nodes. These observations suggest that we should prioritise training data from the last portion, which would lead to overall good LLM generalisation and best accuracy on the end portion, which in turn accelerates search. 

Accordingly, we devise a planner-aware sampling strategy for training data, which prioritises search nodes near the end. In addition, this sampling strategy is general enough to be combined with other coreset selection methods. The proposed strategy incurs, on average, $9.5\%$ fewer A* search steps than uninformed baselines and, in some cases, outperforms models trained with double the amount of data.

Our contributions can be summarised as follows,

\begin{enumerate}
    \item To the best of our knowledge, we are the first work to study the coreset selection problem for A* heuristic learning. Further, we propose a mathematical criterion to select training data based on their distance to goal.
    \item We study the training data requirements for the generalisation of the the learned heuristic function and how heuristic errors affect A* performance, and identify a common requirement shared by the two algorithms. 
    \item Subsequently, we propose a general planner-aware technique to select training data for an LLM-based heuristic function. Our technique outperforms uniform pruning and existing baselines in extensive experiments. 
\end{enumerate}

\section{Related Works}
\label{sec:related-works}
We review several research directions related to our work. For a tabular summary of the works, see Section \ref{sec:rw_summary}. 

\subsection{Learning Heuristics for Planning}

\paragraph{Machine Learning Techniques} Learning for planning problems that aims to reduce the search length can be traced back at least to \citet{yoon2006learning, fern2011first}. This task was posed as a regression problem, learned with neural networks \cite{arfaee2011learning, us2013learning}. Post their success, more recent works explored various neural architectures and objective functions for this problem \cite{chrestien2021heuristic, groshev2018learning, kirilenko2023transpath}. However, existing methods do not cater to specific requirements of the search algorithm.

\paragraph{Search-aware Techniques} Some works consider the requirements of the search algorithm during learning; \citet{yonetani2021path, vlastelica2019differentiation} reformulate each step of the planner as a differentiable function, which can be optimized with the loss calculated at the end of search. However, propagating gradient through time can be compute-intensive. Similarly, \citet{speck2021learning, orseau2023levin, orseau2021policy} learn heuristics by performing reinforcement learning, which could require significant trial-and-error. In this work, we take an alternate data-centric approach to optimize training data. With this, we can lower the computational cost during training, while maintaining the quality of the learned heuristic. 

\subsection{Large Language Models in Search}
\paragraph{Tree Creation by LLMs} In contrast to our focus on LLMs as heuristic functions, previous works have also explored using LLMs as a world model that directly generates the action given the environmental state in search. \citet{yao2024tree} uses such a framework to build a tree and traverses it with depth/breadth-first search, while \citet{hao2023reasoning} extends it to Monte Carlo Tree Search (MCTS), where the LLM selects the tree node to be expanded and generates its children. 

\paragraph{LLMs with External Planners} Besides a heuristic, LLMs have been combined with external planners in various capacities. For instance, \citet{valmeekam2023planning} uses an LLM with the LPG planner \cite{gerevini2002lpg}, which iteratively corrects errors in a plan. Seeding LPG with an LLM plan has been shown to work better than a random plan. LLMs have also been used to translate tasks to formal languages for symbolic solvers like PDDL \cite{liu2023llm+} and ASP \cite{yang2023coupling}. Combining such planners with LLMs has also been explored in dynamic settings to incorporate environment feedback \cite{guan2023leveraging, dagan2023dynamic}. While these works primarily use off-the-shelf LLMs to improve symbolic planners, our work aims to train an LLM.

\paragraph{Improving LLM-based Heuristics}  \citet{shinn2024reflexion} improved LLM heuristics by incorporating failure states into the in-context-learning prompt. This has further been incorporated into tree-based frameworks \cite{zhou2023language}. 
Such failure states are discovered during the course of solving a problem, and thus are restricted to that particular problem instance. 
In contrast, we aim to train a generic heuristic function that works for all problem instances in a domain. An alternate line of work \cite{lehnert2024beyond,gandhi2024stream} utilizes chain-of-thought prompting for LLM planning and trains the LLM on the traces of tree-search algorithms,  implicitly learning an improved heuristic. In contrast, we explicitly learn the heuristic by supervised learning.

\subsection{Optimising Training Data}
\paragraph{Coreset Selection} involves pruning the training dataset to only contain important datapoints, without a significant drop in performance. While various works exist for LLM pre-training \cite{paul2021deep, marion2023less, abbas2023semdedup}, to the best of our knowledge, we are the first work to study this in the context of heuristic learning. Our findings correlate with those of \citet{zhou2023algorithms,sorscher2022beyond}; easier data is required for learning in the low-data regime.

\section{Preliminaries}
\subsection{A* Search}
A* is a tree-based search algorithm that aims to find a path between a start node and any goal node by building a tree $\mathcal{T}$. The algorithm is presented as \autoref{alg:astar}. The set of all tree nodes is denoted as $\mathcal{N}$. For each node $n$, A* search keeps track of two values, (i) historical cost $g(n)$, which is the distance between the start node and $n$ and (ii) heuristic $h(n)$ which is an estimate of the true distance $h^*(n)$ between $n$ and the closest goal node. Each node may be associated with a state $s(n)$. An action modifies the state, causing a transition to a new node. For the search, A* maintains two lists, the frontier list $P_{\text{frt}}$ and the closed list $P_{\text{cld}}$. At the beginning, the closed list is empty and the frontier list is initialised with the start node. The search is terminated when either a goal state is encountered, or $P_{\text{frt}}$ is empty. Each iteration performs two steps, described below.

\paragraph{Selection} This step picks the most-promising leaf node in search tree, which has the least cost $f(n) = g(n) + h(n)$. All leaf nodes are stored in the frontier list. If the state of the selected node is equal to the goal state, the search is terminated. Else, the expansion step is performed. 

\paragraph{Expansion} This step adds new children nodes to the selected node, thereby expanding the search tree. A child node is only added to the search tree if and only if there does not exist a node with the same state in either the frontier, or the closed list, with a lower $f(\cdot)$ value. Finally, the selected node is moved from the frontier to the closed list.

We define the search length $\mathcal{S}$ of A* as the length of the closed list\footnote{which is equal to the number of search iterations} after termination of the search. The use of $h(n)$ makes A* an \textit{informed} search algorithm, significantly reducing the size of the closet list compared to uninformed search. The path from start to goal, defined as $\pi=(n_0,n_1...n_{l})$, is the sequence of $l$ nodes from the start node to the goal node. The start-to-goal path with minimum length is called the \textit{optimal path}, denoted by $\pi^*$. A* guarantees that the resulting path will be optimal if the heuristic is \textit{admissible}, i.e., $h(n) \leq h^*(n), \forall \ n \in \mathcal{N}$. It can be shown that with $h(\cdot) = h^*(\cdot)$ and non-trivial tie-breaking, A* will act as an optimal policy with $\mathcal{S} = |\pi^*|$. An inadmissible heuristic, however, does not necessarily create sub-optimal solutions.

\begin{algorithm}[!t]
\caption{A* Search}
\label{alg:astar}
\begin{algorithmic}
\State $P_{\text{frt}} \gets \{n_{\text{start}}\}$
\State $P_{\text{cld}} \gets \{\}$
\While{$|P_{\text{frt}}| > 0$}
    \State $n \gets \argmin_{n \in P_{\text{frt}}} \ f(n)$ \Comment{Selection}
    \If{$\text{goal-state}(s(n))$}
        \State return $n$
    \EndIf
    \For{$c \in children(n)$} \Comment{Expansion}
        \State $g(c) \gets g(n) + 1$
        \State $f(c) \gets g(c) + h(c)$
        \If{$(\nexists m \in P_{\text{frt}} \cup P_{\text{cld}}, s(c) = s(m))$ \textbf{or}\newline
        \hspace*{4em}$(\exists m \in P_{\text{frt}} \cup P_{\text{cld}}, s(c) = s(m)$ \textbf{and} \newline
        \hspace*{4em} $f(c) < f(m))$\newline\hspace*{2.5em}}
                \State Tree $\mathcal{T} \gets \mathcal{T} \cup \{c\}$
                \State $P_{\text{frt}} \gets P_{\text{frt}} \cup \{c\}$
        \EndIf
    \EndFor
    \State $P_{\text{frt}} \gets P_{\text{frt}} - \{n\}$
    \State $P_{\text{cld}} \gets P_{\text{cld}} \cup \{n\}$
\EndWhile
\end{algorithmic}
\end{algorithm}

\subsection{Training Data for the Heuristic LLM}
Our goal is train a language model $\theta$, that, given a node $n$, can predict the residual $d^*(n) = h^*(n) - h(n)$ between the perfect heuristic $h^*(n)$ and a quick estimate $h(n)$. Given a series of similar problem instances, we derive training data from their A* search trees after a search is complete. For each tree node $n$, computing the ground-truth $d^*(n)$ would require running A* starting from node $n$, which quickly becomes prohibitively expensive as the problem size grows. Following \citet{chrestien2021heuristic,us2013learning}, we only consider nodes on the optimal path. After the first A* run, their $h^*(\cdot)$ is trivial: for any node $n_j \in \pi^*$,\ $h^*(n_j) = |\pi^*| - j$. Formally, the training sequences $\mathcal{X}$ are given by $\mathcal{X} = \bigcup_{\pi^*_i \sim \Gamma_{i=0}^N} \{(n_j, d^*(n_j)), n_j \in \pi^*_i\}$.

\subsection{Loss functions} 
We train the LLM with the L2 loss
\begin{equation}
\mathcal{L}_{L2} = (f_{\theta}(n) - d^*(n))^2
\end{equation}
where $f_{\theta}$ represents a forward pass of the LLM. We use encoder-decoder transformers and add a regression head $\phi_{L2}$ on the decoder that predicts $d^*(n)$ given the $\langle BoS\rangle$ token as the input. 

Additionally, since the LLM can be trained in a text-to-text setting, we train a separate model with the canonical autoregressive loss, given by:
\begin{equation}
\mathcal{L}_{LM} = -\log p(d^*(n)|\theta)
\end{equation}
With $\mathcal{L}_{LM}$, the pre-trained language model head $\phi_{LM}$ is used. 

\subsection{Inference}
Inference involves leveraging the trained LLM in A* search. 
During the expansion step, children nodes to be evaluated are converted into an LLM prompt, from which the LLM predicts $d(n)$. This value is added to the quick estimate of $h(n)$. Notably, only a single forward pass is performed per expansion as we collate all children nodes as one batch. Additionally, we cache these prompts, such that if a state is revisited in another node $m$, $d(m)$ can simply be retrieved. 

For $\theta$ trained with $\mathcal{L}_{LM}$, we perform top-k decoding, with $k = 5$\footnote{This value was arbitrarily chosen and fixed for all experiments. It allows the LLM to make additional choices, without straying too much from the greedy one}, along with self-consistency \cite{wang2022self}, predicting 3 sequences, as this works slightly better in practice.

The exact prompt inputs for the encoder have been provided in Section~\ref{sec:prompts}.

\subsection{Problem Domains} 
We conduct our experiments on three problems domains. Each domain comprises of the in-distribution (IID) and out-of-distribution (OOD) test sets for a total of six datasets. 
\paragraph{Maze Navigation}is a standard maze puzzle that involves finding an unobstructed path from the start to the goal state. The state of a node $s(\cdot)$ is characterized by the position of the player on the board. The quick admissible heuristic function used in the training data (and reference solutions) is the Manhattan distance between the player and the goal positions. Training and validation is performed on sequences derived from mazes of size $20 \times 20$. The IID test split consists of mazes of the same size, while OOD split consists of mazes of size $30 \times 30$. 

\paragraph{Sokoban}is a puzzle game involving a player pushing one or more boxes to fixed docks. This puzzle is considerably harder than maze, since a few wrong moves can lead to deadlocked states. The state of a node is characterized by the position of the player on the board, and the position of the boxes. Note that all boxes and docks are identical. The quick admissible heuristic function used is the sum of the minimum Manhattan distance between the player position and a box, and the sum of Manhattan distances between the boxes and their assigned docks. Boxes are assigned to docks by solving the minimum cost assignment problem with the Hungarian algorithm. Training, validation and IID testing is performed on 2-box problems, while OOD tests are on a mixture of problems with 2, 3 or 4 boxes.

\paragraph{Sliding Tile Puzzle (STP)} is a puzzle consisting of a square board with distinct tiles and one empty space. The task is to move tiles into the empty space to reach a goal configuration. The state of a node is given by the current configuration of the board, and the quick admissible heuristic used is the sum of the Manhattan distance of each tile to its target position. Training, validation and the IID test sets comprise of 3$\times$3 puzzles while the OOD test set consists of harder 4$\times$4 and 5$\times$5 puzzles.

The exact generation and composition of the datasets is described in Section  \ref{sec:data_generation}. In LLM prompts, we use ASCII encoding of the problems shown in Figures \ref{lst:sokoban}, \ref{lst:maze} and \ref{lst:stp}. 

\subsection{Metrics}
\label{sec:metrics}
We adopt several metrics defined by \citet{lehnert2024beyond}, (i) inverse-length-ratio (ILR) to measure the differences in the search length, (ii) success weighted by cost (SWC) to measure the differences in solution length and (iii) optimal \%, to measure the percentage of problems solved optimally.
ILR measures the average inverse ratio between the search length $\tilde{\mathcal{S}}$ of an A* solution, to the optimal reference $\mathcal{S}^*$. It is computed as
\begin{equation}
    ILR = \frac{1}{N}\sum_{i=0}^N \frac{\mathcal{S}^*_i}{\tilde{\mathcal{S}}_i}
\end{equation}
ILR can be averaged over various sets. ILR-on-solved is averaged over all puzzles in the test set and ILR-on-optimal is averaged over all puzzles whose solutions are optimal. Suboptimal solutions, found with inadmissible heuristics, are often discovered before optimal ones, leading to a lower $\mathcal{S}$, but a higher ILR; due to this, ILR-on-optimal allows us to measure the informativeness of the heuristic on equal, minimum length solutions. 

SWC measures the average inverse ratio between the start-to-goal path length $|\tilde{\pi}|$ of an A* solution, to that of an optimal reference, denoted by $|\pi^*|$.
\begin{equation}
    SWC = \frac{1}{N}\sum_{i=0}^N \frac{|\pi^*_i|}{|\tilde{\pi_i}|}
\end{equation}

To measure computational cost, we propose a new metric, \emph{inverse time ratio}, which is defined as the average inverse ratio between the wall-clock time of an A* solution $\tilde{WT}$ and a reference solution $WT^*$, 
\begin{equation}
    ITR = \frac{1}{N}\sum_{i=0}^N \frac{WT^*_i}{\tilde{WT}_i}
\end{equation}

\begin{table}[!t]
\begin{adjustbox}{width=\columnwidth,center}
\begin{tabular}{c|c|cccc}
\midrule
\textbf{Set with \bm{$h^*(\cdot)$}} & \bm{$\sigma$} & \textbf{ILR-on-solved} & \textbf{ILR-on-optimal} & \textbf{SWC} & \textbf{Optimal \%}\\
\midrule
All & - & 2.7356 & 2.7356 & 1.0000 & 100 \\
\midrule

Initial & \multirow{3}{*}{2} & 1.7314 & 1.7717 & 0.9896 & 84.9 \\
Middle &  & 1.8911 & 1.9309 & 0.9908 & 86.4 \\
End &  & \textbf{2.2248} & \textbf{2.2617} & \textbf{0.9919} & \textbf{87.9} \\
\midrule
Initial & \multirow{3}{*}{4} & 1.0842 & 1.1912 & \textbf{0.9530} & 46.1 \\
Middle &  & 1.1604 & 1.2924 & 0.9516 & 46.2 \\
End &  & \textbf{1.5439} & \textbf{1.7389} & 0.9520 & \textbf{46.3} \\
\midrule
Initial & \multirow{3}{*}{6} & 0.8579 & 0.9827 & 0.9229 & 28.6 \\
Middle &  & 0.9192 & 1.0811 & \textbf{0.9232} & \textbf{29.3} \\
End &  & \textbf{1.2157} & \textbf{1.5287} & 0.9202 & 28.1 \\
\bottomrule
\end{tabular}
\end{adjustbox}
\caption{Experimental results with the oracle heuristic on the validation puzzles of maze navigation.}
\label{tbl:oracle_heuristic}
\end{table}

\section{Disentangling A* and Heuristic Learning}
\begin{figure*}[htbp]
    \centering
    \begin{subfigure}{0.46\textwidth}
        \centering
        \includegraphics[width=\textwidth]{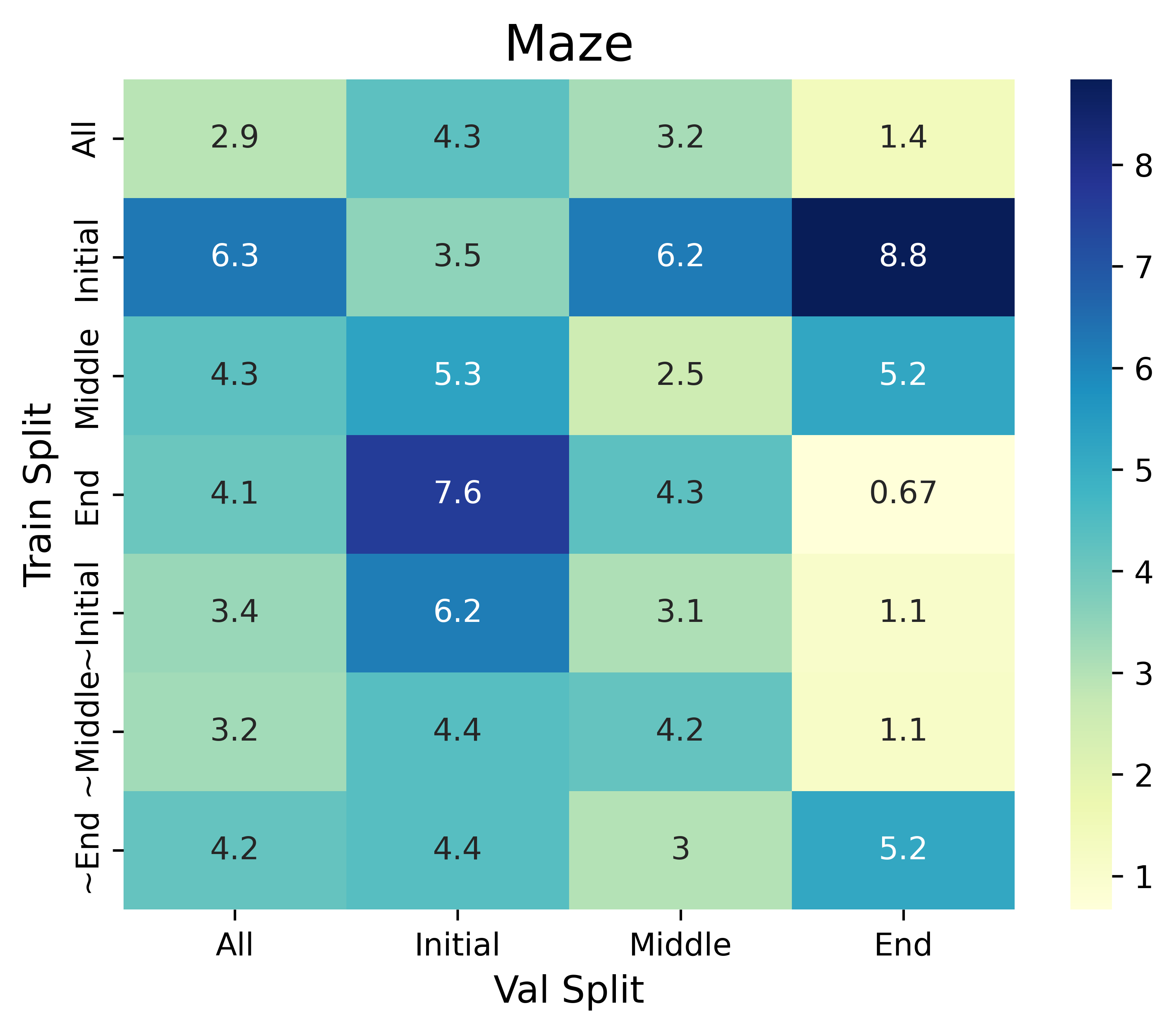}
    \end{subfigure}
    \hfill
    \begin{subfigure}{0.46\textwidth}
        \centering
        \includegraphics[width=\textwidth]{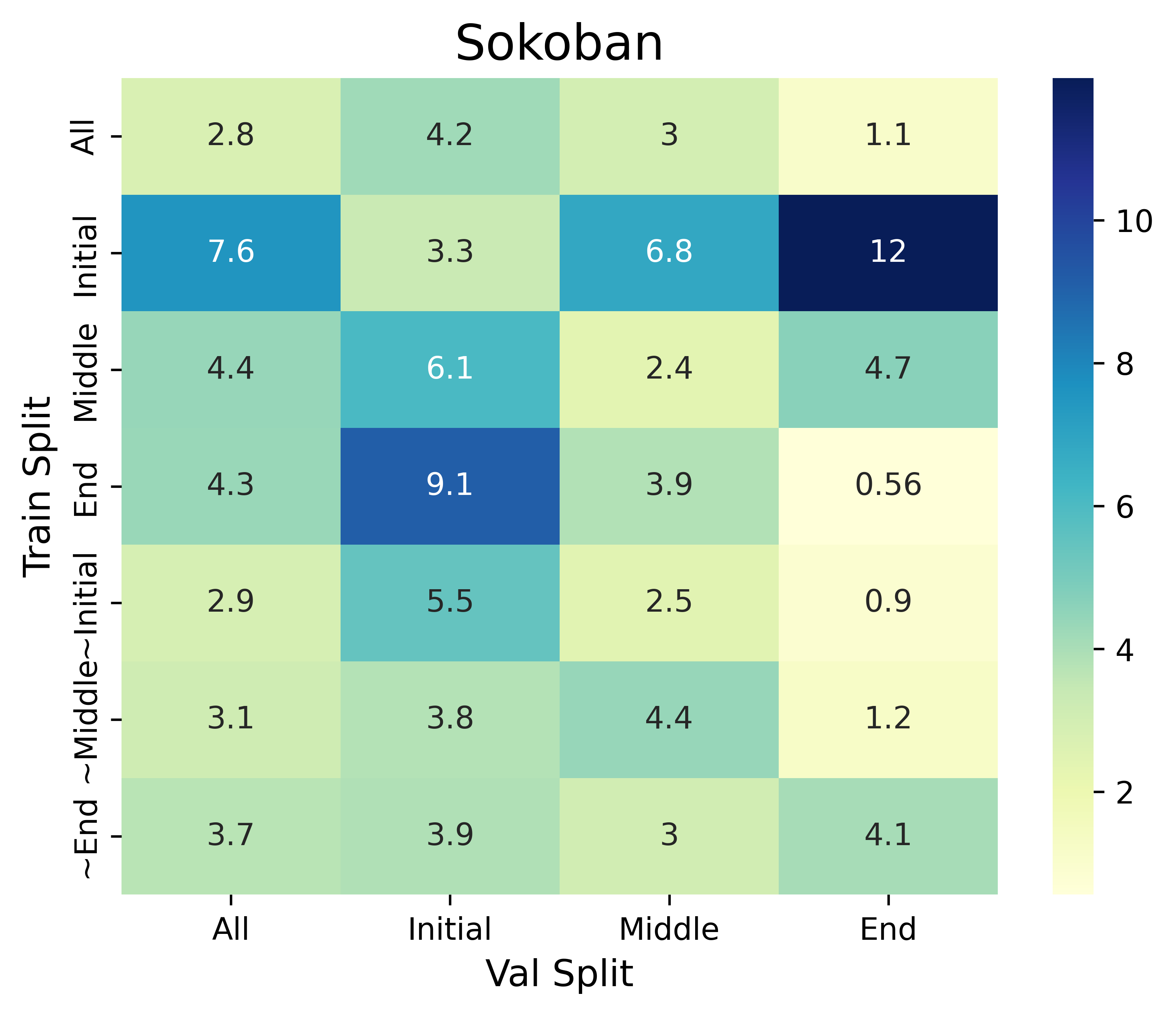}
    \end{subfigure}
    \caption{Validation MAE of models trained on the \textit{Initial}, \textit{Middle}, \textit{End}, and \textit{All} splits, and their corresponding exclusion sets. A lower value shows better generalisation.}
    \label{tbl:generalisation}
\end{figure*}

\begin{table*}[!t]
\begin{adjustbox}{width=\textwidth,center}
\begin{tabular}{c|c|cccc|cccc}
\toprule[1pt]
\multicolumn{2}{c}{\textbf{Test Splits} $\rightarrow$} & \multicolumn{4}{|c}{\textbf{IID}} & \multicolumn{4}{|c}{\textbf{OOD}}\\
\midrule
\textbf{Train Split} & \textbf{Domain} & \textbf{ILR-on-solved} & \textbf{ILR-on-optimal} & \textbf{SWC} & \textbf{Optimal \%} & \textbf{ILR-on-solved} & \textbf{ILR-on-optimal} & \textbf{SWC} & \textbf{Optimal \%}\\
\midrule[1pt]
All & \multirow{6}{*}{Maze} & 1.5666 & 1.5654 & 0.9972 & 97.60 & 1.3320 & 1.3309 & 0.9965 & 96.00\\
\cmidrule{1-1} \cmidrule{3-10}
Initial &  & 0.9101 & 0.9101 & \textbf{1.0000} & \textbf{100.0} & 0.8193 & 0.8193 & \textbf{1.0000} & \textbf{100.0} \\
Middle &  & 0.8370 & 0.837 & \textbf{1.0000} & \textbf{100.0} & 0.8059 & 0.8059 & \textbf{1.0000} & \textbf{100.0}\\
End &  & \textbf{1.2081} & \textbf{1.2033} & 0.9974 & 97.40 & \textbf{1.1018} & \textbf{1.1055} & 0.9957 & 95.40 \\
\cmidrule{1-1} \cmidrule{3-10}
$\sim$ Initial &  & 1.2117 & 1.2132 & 0.9989 & 99.00 & 1.0581 & 1.0594 & 0.9992 & 98.80\\
$\sim$ Middle &  & \textbf{1.6053} & \textbf{1.6151} & 0.9907 & 92.80 & \textbf{1.2476} & \textbf{1.2360} & 0.9950 & 94.40\\
$\sim$ End &  & 0.9202 & 0.9202 & \textbf{1.0000} & \textbf{100.0} & 0.9198 & 0.9198 & \textbf{1.0000} & \textbf{100.0} \\
\midrule[1pt]

All & \multirow{6}{*}{Sokoban} & 8.3800 & 8.8785 & 0.9761 & 73.94 & 11.1967 & 11.7906 & 0.9815 & 74.46\\
\cmidrule{1-1} \cmidrule{3-10}
Initial &  & 0.6658 & 0.6661 & \textbf{0.9967} &\textbf{ 93.66} & 0.5940 & 0.5917 & 0.9956 & 90.12\\
Middle &  & 0.9710 & 1.0049 & 0.9901 & 83.80 & 0.8148 & 0.8399 & 0.9904 & 84.34\\
End &  & \textbf{3.0312} & \textbf{3.0642} & 0.9965 & \textbf{93.66} & \textbf{2.7465} & \textbf{2.7721} & \textbf{0.9986} & \textbf{96.39}\\
\cmidrule{1-1} \cmidrule{3-10}
$\sim$ Initial & & 6.1912 & 6.5422 & \textbf{0.9862} & \textbf{82.04} & 9.2832 & 9.8333 & \textbf{0.9893} & \textbf{83.86} \\
$\sim$ Middle &  & \textbf{9.7389} & \textbf{9.9559} & 0.9578 & 56.69  & \textbf{16.3567} & \textbf{18.1764} & 0.9650 & 61.45 \\
$\sim$ End &  & 2.8397 & 2.9638 & 0.9854 & 80.28 & 2.9484 & 3.0910 & 0.9854 & 78.07\\
\bottomrule[1pt]

\end{tabular}
\end{adjustbox}
\caption{Results from LLM heuristics trained on different data splits, demonstrating the importance of the \textit{End} split for generalisation to A* search on both maze and Sokoban.}
\label{tbl:trained_heuristic}
\end{table*}

\label{sec:disentangling}
\subsection{Understanding Requirements of A*}
\label{sec:a-star-requirements}
Prediction errors by the LLM in the learned heuristic function are inevitable. In this section, we aim to examine two research questions: (i) how the prediction errors in the learned heuristic function affects the search length $\mathcal{S}$, and (ii) how they affect optimality of the solutions.

Specifically, we start with the oracle heuristic $h^*(n)$ and artificially introduce error in different sections of the search trajectory in order to observe effects on $\mathcal{S}$ and optimality. The search tree is divided into three sets---\textit{initial}, \textit{middle} and \textit{end}. A node $n$ is placed in the initial set if its cost places itself in the first third of the optimal path: $g(n) < |\pi^*|/3$. Alternatively, it may be placed in the middle set if $|\pi^*|/3 \leq g(n) < 2|\pi^*|/3$, and in the end set if $g(n) \geq 2|\pi^*|/3$. 
We introduce zero-mean Gaussian error by drawing a random value from $\mathcal{N}(0, \sigma)$ and adding it to $h^*(n)$. In each experiment, we introduce errors in two of three sections and use the oracle in one section. We use maze as the domain of experiment and obtain the oracle heuristic $h^*(\cdot)$ by running Dijkstra's algorithm on the maze, starting from the goal.

\paragraph{Results} The results are shown in \autoref{tbl:oracle_heuristic}. The rows \emph{All}, \emph{Initial}, \emph{Middle}, and \emph{End} indicate the tree section where the oracle is utilized, and \emph{All} means the oracle is always used. Clearly, the oracle heuristic gives the best performance, but that is not easy to achieve by a learned model. Amongst other experiment conditions, with the same $\sigma$, using $h^*(\cdot)$ on nodes in the end set performs the best on both ILR-on-solved and ILR-on-optimal. Moreover, the absolute differences in performance by using $h^*(\cdot)$ in the middle and end sets are larger than the differences between middle and initial. These performance gaps are larger with a higher $\sigma$. 

There does not seem to be a clear trend between SWC and Optimal \% amongst the three sets. Both metrics go down with increasing $\sigma$. This is not surprising, since with higher error, the heuristic will be inadmissible more frequently, thereby increasing the probability of finding longer, suboptimal solutions.

\paragraph{Implications} The most important implication of these experiments is that, if we can only minimize errors of the heuristic function on one section of the search trajectory, we should choose the end section, which is closest to the goal. Doing so yields the highest ILR. Speculatively, erroneous decisions earlier in the trajectory may be corrected later, if we can make good decisions near the end of the search process.  

\subsection{Understanding Generalisation of Heuristic Learning}
In this section, we explore how training on pairs of (node, distance-to-goal) affects the generalisation of the heuristic-learning LLM. We create four training splits by uniformly sampling nodes on the optimal path from the \textit{Initial}, \textit{Middle}, and \textit{End} sections of the path. The \textit{All} set contains nodes uniformly sampled from all three sections. 
Additionally, we also create exclusion sets, which excludes one of the three sections, and these sets are denoted as \textit{$\sim$Initial}, \textit{$\sim$Middle} and \textit{$\sim$End}. For instance, \textit{$\sim$Initial} contains only data sampled from the \textit{Middle} and \textit{End} sets. All training splits have the same size. 

We adopt the following evaluation metrics: (i) mean absolute error (MAE) on validation splits containing nodes from each of the aforementioned splits, and (ii) ILR achieved by applying the trained models as heuristic functions for A*. While (i) directly evaluates the generalisation of the model, (ii) provides a more realistic test of how well the trained model works with A*. 

Each training split contains 12k and 8k nodes in total for maze and Sokoban, respectively. All models are initialized with code-t5-small. Hyperparameter details are mentioned in Section~\ref{sec:hyperparameters}.

\paragraph{Results} The LLM generalisation results are shown in \autoref{tbl:generalisation} and results from applying different LLMs with A* are shown in \autoref{tbl:trained_heuristic}. First, as we expect, each split generalises the best to itself, but shows poor generalisation to the others. \emph{All} achieves the best generalisation to each split. Second, on ILR, \textit{End} performs the best when combined with A*. However, this is still inferior to the performance of \textit{All}. This is consistent with the trends observed in Section \ref{sec:a-star-requirements}.

Amongst the exclusion sets, we observe that $\sim$\textit{End} achieves the worst generalisation and the worst ILR in both domains and both IID and OOD test splits. The comparison between the other two sets is mixed. $\sim$\textit{Middle} has the best ILR performance, whereas $\sim$\textit{Initial} performs well on Optimal $\%$ and SWC.

\paragraph{Implications} Heuristics learned from the end set performs the best on MAE and well on ILR, showing that we need the end set in the training mix. These nodes can be considered easier than others because it is easy to foresee the distance to goal for a node positioned near the goal. However, the good performance of $\sim$\textit{Middle} and $\sim$\textit{Initial} suggests that easy nodes by themselves are not enough, and we should expose the model to some difficult nodes from the other sets, which are further away from the goal.  

\begin{table*}[!t]
\begin{adjustbox}{width=\textwidth,center}
\begin{tabular}{c|c|cccc|cccc}
\midrule
\multicolumn{2}{c}{\textbf{Test Splits} $\rightarrow$} & \multicolumn{4}{|c}{\textbf{IID}} & \multicolumn{4}{|c}{\textbf{OOD}}\\
\midrule
\textbf{Train Split} & \textbf{Domain} & \textbf{ILR-on-solved} & \textbf{ILR-on-optimal} & \textbf{SWC} & \textbf{Optimal \%} & \textbf{ILR-on-solved} & \textbf{ILR-on-optimal} & \textbf{SWC} & \textbf{Optimal \%}\\
\midrule

Full-data & \multirow{5}{*}{Maze} & 1.6739 & 1.6756 & 0.9967 & 97.0 & 1.2755 & 1.2730 & 0.9967 & 95.8\\
\cline{1-1} \cline{3-10}
$\mathcal{X} \sim \mathcal{U}(n)$ & & 1.5666 & 1.5654 & \textbf{0.9972} & \textbf{97.6} & 1.3320 & 1.3309 & 0.9965 & 96.0\\
$\mathcal{X} \sim \mathcal{D}(n, 2)$ & & 1.7029 & 1.7035 & 0.9958 & 96.6 & 1.3365 & 1.3354 & 0.9964 & 95.0 \\
$\mathcal{X} \sim SD$ & & 1.6412 & 1.6453 & 0.9941 & 95.2 & 1.2823 & 1.2821 & \textbf{0.9980} & \textbf{97.6} \\
$\mathcal{X} \sim SD + \mathcal{D}(n, 2)$ & & \textbf{1.7182} & \textbf{1.7245} & 0.9927 & 94.6 & \textbf{1.3568} & \textbf{1.3521} & 0.9968 & 96.4 \\
\midrule

Full-data & \multirow{5}{*}{Sokoban} & 11.6416 & 12.5933 & 0.9834 & 79.93 & 14.7093 & 15.2655 & 0.9847 & 77.83\\
\cline{1-1} \cline{3-10}
$\mathcal{X} \sim \mathcal{U}(n)$ & & 8.3800 & 8.8785 & 0.9761 & 73.94 & 11.1967 & 11.7906 & 0.9815 & \textbf{74.46}\\
$\mathcal{X} \sim \mathcal{D}(n, 0.8)$ & & 10.2077 & 10.8168 & \textbf{0.9808} & \textbf{75.70} & 13.7706 & 13.7546 & \textbf{0.9828} & 77.11\\
$\mathcal{X} \sim SD$ & & 10.8579 & 11.5282 & 0.9702 & 68.66 & 14.9133 & 15.4475 & 0.9757 & 71.58 \\
$\mathcal{X} \sim SD + \mathcal{D}(n, 5)$ & & \textbf{11.5184} & \textbf{11.8487} & 0.9732 & 68.66 & \textbf{15.8553} & \textbf{15.9748} & 0.9772 & 72.05 \\
\midrule

Full-data & \multirow{5}{*}{STP} & 4.1509 & 4.5750 & 0.9806 & 77.4 & 1.5012 & 1.5374 & 0.9860 & 84.4 \\
\cline{1-1} \cline{3-10}
$\mathcal{X} \sim \mathcal{U}(n)$ & & 3.4040 & 3.7777 & 0.9755 & 72.8 & 1.3054 & 1.3789 & 0.9859 & 85.2 \\
$\mathcal{X} \sim \mathcal{D}(n, 5)$ & & 3.4758 & 3.9686 & \textbf{0.9765} & \textbf{73.8} & 1.4265 & 1.4606 & \textbf{0.9946} & \textbf{93.0} \\
$\mathcal{X} \sim SD$ & & 3.5372 & 4.2400 & 0.9617 & 60.6 & \textbf{2.4353} & \textbf{2.7080} & 0.9804 & 77.4 \\
$\mathcal{X} \sim SD + \mathcal{D}(n, 5)$ & & \textbf{4.2779} & \textbf{4.7384} & 0.9723 & 70.6 & 1.7050 & 1.8955 & 0.9694 & 69.6 \\
\bottomrule

\end{tabular}
\end{adjustbox}
\caption{Experimental results with $\mathcal{L}_{L2}$ by sampling from the $\mathcal{D}(n, \tau)$ distribution. Best scores are in \textbf{bold}.}
\label{tbl:importance_sampling}
\end{table*}

\section{Proposed Solution}
\paragraph{The Utility of a Node in Accelerating Search}
Inspired by experiments in Section~\label{sec:disentangling}, we propose to quantify the utility of a node in reducing the search length as,
\begin{equation}
    \label{eqn: contr}
    \mathcal{C}(n) = \log \left(\frac{|\pi^*|}{|\pi^*| - g(n)}\right)
\end{equation}
$\mathcal{C}(\cdot)$ assigns higher values to nodes closer to the goal.

While there can be nodes with $g(n) \geq |\pi^*|$, since they are never added to the tree, $\mathcal{C}(\cdot)$ is not defined for them. Considerations and other choices for $\mathcal{C}(\cdot)$ are discussed in Section~\ref{sec:ablations}.

\begin{table*}[!t]
\begin{adjustbox}{width=\textwidth,center}
\begin{tabular}{c|c|c|cccc|cccc}
\toprule
\multicolumn{3}{c|}{\textbf{Test Splits} $\rightarrow$} & \multicolumn{4}{c|}{\textbf{IID}} & \multicolumn{4}{c}{\textbf{OOD}} \\
\midrule
\textbf{Base Model} & \textbf{Train Split} & \textbf{Domain} & \textbf{ILR-on-solved} & \textbf{ILR-on-optimal} & \textbf{SWC} & \textbf{Optimal \%} & \textbf{ILR-on-solved} & \textbf{ILR-on-optimal} & \textbf{SWC} & \textbf{Optimal \%}\\
\midrule
\multirow{2}{*}{codet5-base} & $\mathcal{X} \sim \mathcal{U}(n)$ & \multirow{6}{*}{Maze} & 1.7218 & 1.7245 & \textbf{0.9965} & \textbf{97.0} & 1.2841 & 1.2722 & 0.9970 & 96.4 \\
 & $\mathcal{X} \sim \mathcal{D}(n, 2)$ &  & \textbf{1.8112} & \textbf{1.8142} & 0.9957 & 96.8 & \textbf{1.3460} & \textbf{1.3422} & \textbf{0.9977} & \textbf{97.2} \\
\cline{1-2}\cline{4-11}
\multirow{2}{*}{codet5-large} & $\mathcal{X} \sim \mathcal{U}(n)$ &  & 1.2963 & 1.2966 & \textbf{0.9995} & \textbf{99.6} & 1.1531 & 1.153 & \textbf{0.9994} & \textbf{99.2} \\
 & $\mathcal{X} \sim \mathcal{D}(n, 1)$ &  & \textbf{1.6920} & \textbf{1.6982} & 0.9964 & 97.4 & \textbf{1.3101} & \textbf{1.3088} & 0.9980 & 97.6 \\
\cline{1-2}\cline{4-11}
\multirow{2}{*}{t5-small} & $\mathcal{X} \sim \mathcal{U}(n)$ &  & 1.5447 & 1.5483 & \textbf{0.9967} & \textbf{97.2} & 1.3287 & 1.3276 & \textbf{0.9975} & \textbf{97.0}\\
 & $\mathcal{X} \sim \mathcal{D}(n, 2)$ & & \textbf{1.5785} & \textbf{1.5818} & 0.9957 & 96.4 & \textbf{1.3404} & \textbf{1.3378} & 0.9974 & \textbf{97.0}\\
\midrule
\multirow{2}{*}{codet5-base} & $\mathcal{X} \sim \mathcal{U}(n)$ & \multirow{6}{*}{Sokoban} & \textbf{10.8858} & 11.1579 & 0.9770 & 71.83 & 14.4553 & 14.4831 & 0.9810 & 74.70 \\
 & $\mathcal{X} \sim \mathcal{D}(n, 2)$ &  & 10.6828 & \textbf{11.1692} & \textbf{0.9791} & \textbf{73.94} & \textbf{15.0611} & \textbf{15.2904} & \textbf{0.9828} & \textbf{76.39} \\
\cline{1-2}\cline{4-11}
\multirow{2}{*}{codet5-large} & $\mathcal{X} \sim \mathcal{U}(n)$ &  & 10.3732 & \textbf{10.7997} & 0.9788 & 74.3 & 12.8759 & 12.9480 & 0.9830 & 76.39 \\
 & $\mathcal{X} \sim \mathcal{D}(n, 2)$ &  & \textbf{10.3778} & 10.7343 & \textbf{0.9850} & \textbf{80.99} & \textbf{13.0179} & \textbf{12.9534} & \textbf{0.9891} & \textbf{83.37} \\
 \cline{1-2}\cline{4-11}
\multirow{2}{*}{t5-small} & $\mathcal{X} \sim \mathcal{U}(n)$ &  &  10.8294 & \textbf{11.1671} & 0.9707 & 70.07 & 11.4536 & 11.2696 & \textbf{0.9882} & \textbf{80.96}\\
 & $\mathcal{X} \sim \mathcal{D}(n, 2)$ &  & \textbf{10.9260} & 10.9835 & \textbf{0.9803} & \textbf{75.00} & \textbf{12.4921} & \textbf{12.7784} & 0.9865 & 78.80\\

\bottomrule
\end{tabular}
\end{adjustbox}
\caption{Experiments with $\mathcal{L}_{L2}$, showing the effects of planner-aware sampling on various models.}
\label{tbl:model_scale}
\end{table*}

\paragraph{Planner-aware Sampling} We have shown that accurate prediction of the heuristic for nodes near the goal will lead to maximal reduction of the search length. Additionally, we want to include nodes from the initial and middle sets as well, to optimize ILR performance. Thus, we propose to sample from a distribution $\mathcal{D}(\cdot)$ that prioritises these nodes, based on \autoref{eqn: contr} (as opposed to a uniform distribution), given by,
\begin{equation}
\mathcal{D}(n, \tau) = SoftMax\left(\frac{1}{\tau}\mathcal{C}(n)\right), \forall n \in \pi^*
\end{equation}
where $\tau$ denotes temperature. Increasing $\tau$ increases the hardness of the training dataset, thereby increasing the number of nodes sampled from the initial and middle sets.

\begin{algorithm}[t]
\caption{Combining planner-aware sampling with a coreset selection baseline $\Psi$.}
\label{alg:pab}
\begin{algorithmic}
\State Assume $m$ nodes are sampled from a problem,
\State $\mathcal{S}_1 \gets \{n_i \sim \Psi(n)\ |\ i \in [1, m]\}$
\State $\mathcal{S}_2 \gets \{n_i \sim \mathcal{D}(n, \tau)\ |\ i \in [1, m]\}$
\State $\mathcal{P}(n_i) \gets 
\left\{\begin{array}{ll}
      \frac{2}{|\mathcal{S}_1| + |\mathcal{S}_2|}, & n_i \in \mathcal{S}_1 \cap \mathcal{S}_2 \\
      \frac{1}{|\mathcal{S}_1| + |\mathcal{S}_2|}, & otherwise \\
\end{array} \right.$
\State $\mathcal{X} \gets \{n_i \sim \mathcal{P}(\mathcal{S}_1 \cup \mathcal{S}_2))\ |\ i \in [1, m]\}$

\end{algorithmic}
\end{algorithm}

\paragraph{Combining with Baselines} Planner-aware sampling can be easily combined with any coreset selection baseline to enhance it for this task. This is done by first sampling two sets of nodes (without replacement), once using any coreset selection baseline $\Psi$, and another with $\mathcal{D}(n, \tau)$. Post this, the nodes can be resampled, without replacement, from the union of these two sets, where nodes appearing in both the sets are twice as likely to get sampled than those appearing in only a single set. This procedure is summarised in \autoref{alg:pab}.

\section{Experiments}

\paragraph{Baselines} The proposed sampling method is denoted as $\mathcal{D}(n, \tau)$. The \textit{full-data} baseline trains on all nodes (22.3k nodes for maze, 26.3k for Sokoban and 23.7k for STP) on the optimal path without subsampling. The uniform sampling baseline $\mathcal{U}(n)$ gives equal probability to all nodes. To the best of our knowledge, there are no search-aware coreset selection methods. Hence, we adopt as a baseline an LLM-based coreset selection method, SemDeDup ($SD$)~\cite{abbas2023semdedup}, which discards semantically similar data points from the training dataset. On top of $SD$, we apply \autoref{alg:pab} to make it search-aware ($SD + \mathcal{D}(n, \tau)$). All coreset selection methods select 8k nodes for STP, and as before, 8k for Sokoban and 12k for maze.

\paragraph{Results} The results using the $\mathcal{L}_{L2}$ loss are shown in \autoref{tbl:importance_sampling}. We defer results with $\mathcal{L}_{LM}$ to \autoref{tbl:importance_sampling_lm}.
$\mathcal{D}(n, \tau)$ consistently outperforms uniform sampling on ILR by an average of $4.4\%$ on maze, $5.7\%$ on STP, and a much larger margin of $12.5\%$ on Sokoban. On maze, $\mathcal{D}(n, \tau)$ also outperforms the full-data baseline on OOD data, which is trained on $46.5\%$ more data points. These results also extend to $\mathcal{L}_{LM}$, where $\mathcal{D}(n, \tau)$ outperforms $\mathcal{U}(n)$ by an average of $5\%$.

In terms of metrics of solution optimality (SWC and Optimal \%), $\mathcal{D}(n, \tau)$ remains competitive and is better than the baselines by an average of $0.24\%$. Interestingly, training on all the data gives higher performance on optimality metrics, which could be a consequence of lower validation error, due to more training data.

Notably, the $SD$ coreset selection baseline, developed for LLMs, also performs quite well. However, $SD$ augmented with $\mathcal{D}(n, \tau)$ outperforms all other methods, except on STP OOD, by an average of $8.75\%$.

\paragraph{Model Scale and Pre-training} To test the effectiveness of our method while scaling up the LLM, we experiment with thre LLMs, t5-small (60M), codet5-base (220M), and codet5-large (770M). \autoref{tbl:model_scale} demonstrates similar trends  of $\mathcal{D}(n, \tau)$ outperforming $\mathcal{U}(n)$.   Interestingly, the performance of larger models is not always better than that of smaller models. This could be attributed to the fact that our experiments have been performed in the low-data regime and large models cause overfitting. Studying the effects of scaling up data with parameters is left for future works. The learned heuristics with larger models are more optimal, suggesting less error in the predictions. 

\paragraph{Time Cost of LLM Inference}
It is well accepted in the planning domain that a more informative heuristic is more expensive to compute \cite{bylander1994computational}. While LLMs incur additional time during inference, the learned heuristic is informative enough to amortize the extra time cost. We use ITR as the evaluation metric, which shows speed-ups in wall-clock search time compared to the LLM-free A* search. An ITR value $>$ 1 implies that the LLM heuristic is faster than the base heuristic. 

Experiments are performed on the $\mathcal{D}(n, \tau)$ models (from \autoref{tbl:importance_sampling}), trained with $\mathcal{D}(n, \tau)$ sampling. We show the results in \autoref{tbl:computational_cost}. Due to its difficulty, Sokoban has a high number of explored nodes in each problem (often greater than 10k). With the LLM heuristic, the ITR on the most difficult OOD test split is greater than one. On the IID set, with easier problems of shorter search lengths, the ITR is close to one, but does not surpass it. Similarly, the ITR is less than one on maze, which consists of easier problems with low $\mathcal{S}$. Since the number of nodes is already quite low (mostly between 2k and 2.5k), a reduction does not necessarily bring about wall-clock speed-up. With this, we conclude that the LLM search heuristic is the most beneficial on hard OOD problems, which is also where direct inference from LLMs struggle the most \citep{wu2024reasoning} and exactly what is needed for boostrapping from easy to hard problems.

Interestingly, $\mathcal{L}_{LM}$ is almost $2.5\times$ slower on average than $\mathcal{L}_{L2}$, despite the ILR being only $1.1\times$ worse. This suggests that though $\phi_{LM}$ is capable of learning an informative heuristic, the forward pass through the larger linear layer, along with stochastic decoding, negatively affects efficiency.

\begin{table}[!t]
\begin{adjustbox}{width=0.95\columnwidth,center}
\begin{tabular}{c|c|cc}
\midrule
\textbf{Domain} & \textbf{Test Split}& \textbf{ITR-on-solved} & \textbf{ITR-on-optimal}\\
\midrule
\multicolumn{4}{c}{\textbf{Model :} \bm{$\mathcal{L}_{L2}$}}\\
\midrule
\multirow{2}{*}{Sokoban} & IID & 0.8167 & 0.8735\\
                         & OOD & 5.9441 & 5.9215\\
                         \midrule
\multirow{2}{*}{Maze} & IID & 5.122$e-3$ & 5.127$e-3$\\
                         & OOD & 5.062$e-3$ & 5.079$e-3$\\
                         \midrule
\multicolumn{4}{c}{\textbf{Model :} \bm{$\mathcal{L}_{LM}$}}\\
                         \midrule
\multirow{2}{*}{Sokoban} & IID & 0.2626 & 0.2611\\
                         & OOD & 2.7250 & 2.3978 \\
                         \midrule
\multirow{2}{*}{Maze} & IID & 1.958$e-3$ & 1.963$e-3$\\
                         & OOD & 2.090$e-3$ & 2.096$e-3$\\
\bottomrule
\end{tabular}
\end{adjustbox}
\caption{Speed-ups in wall-clock search time achieved by using the trained language model as a heuristic.}
\label{tbl:computational_cost}
\end{table}

\paragraph{Training Target}
Between $\mathcal{L}_{L2}$ and $\mathcal{L}_{LM}$ models, the former consistently outperforms the latter on the IID test split, while on OOD, the results are mixed, with $\mathcal{L}_{LM}$ being slightly better, at least with uniform sampling on Sokoban. $\mathcal{L}_{LM}$ is more aligned with the pre-training of the base model, which may have improved generalisation beyond the training data distribution.

Another interesting observation is that the best hyperparameter $\tau$ used with $\mathcal{D}(n,\tau)$, tuned on the validation set, is usually higher for $\mathcal{L}_{LM}$, suggesting that it has a higher preference for data points in the initial set, which can be considered harder than other nodes.

\section{Conclusion}
In this work, we study the training data requirements to learn a strong heuristic for A* search. We find that accurate prediction of heuristics for nodes close to the goal are the most important for A* speed. Similarly, generalization of the LLM heuristic requires training on nodes near the goal. Based on these insights, we propose a mathematical formula to select search nodes as training data. This results in substantially reduced search lengths and significant wall-clock speedups on hard problems. Our study lays the groundwork for bootstrapped heuristic learning, which learns heuristic functions for increasingly larger problems using solved problems of smaller sizes. Referred to as the data flywheel, such techniques hold promise to scale up the capabilities of LLM + tree search\footnote{\url{https://twitter.com/DrJimFan/status/1834279865933332752}}. 

\section*{Acknowledgments}
We gratefully acknowledge the support by the Nanyang Associate Professorship, the National Research Foundation Fellowship (NRF-NRFF13-2021-0006), Singapore, and the Alibaba-NTU Global e-Sustainability CorpLab (ANGEL) under Project I2301E0026. Any opinions, findings, conclusions, or
recommendations expressed in this material are
those of the authors and do not reflect the views of
 the funding agencies.
\section*{Limitations}
Our study is restricted to classical puzzle domains, maze, Sokoban and STP. While we expect our domain-independent analysis to generalise to other problems, experimental verification would be necessary to verify that conjecture. Moreover, since our work focuses on language models used as heuristics, it inherits the bias and fairness concerns associated with language models, which should be taken into consideration when deploying such models.

\bibliography{custom}
\appendix
\section{Appendix}
\subsection{Data Generation}
\label{sec:data_generation}
\paragraph{Maze} We generate mazes with a modified Prim's algorithm\footnote{\url{https://github.com/john-science/mazelib}}. The start and goal states are randomly chosen until the following criteria are met, (i) length of the optimal plan > $O_l$, (ii) ratio between length of closed set after search and length of optimal plan is > $\alpha = 3.5$. If either of these are not met within 10 tries, a new maze is generated. Criterion (i) ensures that the start and goal positions are not too close and (ii) ensures that there are sufficient number of additional expanded nodes. It serves as a surrogate for the measure of hardness $h*(n_s)/h(n_s)$ where $n_s$ is the start node, proposed in \citet{takahashi2019learning}. The surrogate is used since it is more aligned with the chosen metrics (ILR) in this work. However, this method only creates a maze with a single path to the goal. To get multiple paths, each node is designated to either be closer to the start, or to the goal, and walls are randomly broken at the boundary of these groups\footnote{\url{https://stackoverflow.com/a/22308159}}.
\paragraph{Sokoban} This dataset is adapted from the open-source boxoban dataset, proposed in \citet{guez2019investigation}. For the training puzzles, we randomly shuffle the provided training set from the "unfiltered" split, followed by subsampling $B$ boxes per puzzle. We use the same filters as maze, but with different hyperparameters. The IID test split uses the same criteria, but samples puzzles from the testing set of boxoban. To reduce the data creation time, we constrain the number of iterations required by A* to solve a puzzle between $\beta_{min}$ and $\beta_{max}$. The OOD split is curated to contain a mix of harder puzzles with varying number of boxes, length of optimal plans and higher number of iterations. All puzzles have size $10\times10$, $O_l = 20$ and $\alpha = 6$.

\paragraph{STP} We generate 3$\times$3 puzzles by randomly generating a sequence of tiles, checking if it is solvable with A*. For puzzles with a width greater than 3, we start from the goal configuration and perform 20 - 30 random moves to scramble the puzzle, from the goal state. For all puzzles, $\alpha = 6$, $O_l = 20$, $\beta_{min} = 0$ and $\beta_{max} = 5k$. To keep the symbols in the puzzle uniform between training and inference, the generation of puzzles is done with digits, however, they are fed to the model as alphabets. For each puzzle, we uniformly sample without replacement the required number of alphabets, sort them alphabetically and assign them to the digits.

The exact statistics are in \autoref{tbl:datasets_maze}, \autoref{tbl:datasets_Sokoban} and \autoref{tbl:datasets_stp}.
\begin{table}[ht]
\begin{adjustbox}{width=0.8\columnwidth,center}
\begin{tabular}{c|cccc}
\midrule
\textbf{Split} & \textbf{\# puzzles} & \textbf{Size} & \bm{$O_l$}\\
\midrule
Train & 750 & $20\times20$ & 20\\
Val & 750 & $20\times20$ & 20\\
Test IID & 500 & $20\times20$ & 20\\
Test OOD & 500 & $30\times30$ & 30\\
\bottomrule
\end{tabular}
\end{adjustbox}
\caption{Dataset statistics for maze.}
\label{tbl:datasets_maze}
\end{table}

\begin{table}[ht]
\begin{adjustbox}{width=0.8\columnwidth,center}
\begin{tabular}{c|cccc}
\midrule
\textbf{Split} & \textbf{\# puzzles} & \bm{$B$} & \bm{$\beta_{max}$} & \bm{$\beta_{min}$}\\
\midrule
Train & 1000 & 2 & 7k & 0\\
Val & 1000 & 2 & 7k & 0\\
Test IID & 284 & 2 & 7k & 0\\
\midrule
\multirow{5}{*}{Test OOD} & 15 & 2 & 14k & 7k \\
& 100 & 3 & 7k & 0 \\
& 100 & 3 & 14k & 7k \\
& 100 & 4 & 7k & 0 \\
& 100 & 4 & 14k & 7k \\
\bottomrule
\end{tabular}
\end{adjustbox}
\caption{Dataset statistics for Sokoban.}
\label{tbl:datasets_Sokoban}
\end{table}

\begin{table}[ht]
\begin{adjustbox}{width=0.6\columnwidth,center}
\begin{tabular}{c|ccc}
\midrule
\textbf{Split} & \textbf{\# puzzles} & \textbf{Size}\\
\midrule
Train & 1000 & $3\times3$\\
Val & 1000 & $3\times3$\\
Test IID & 500 & $3\times3$\\
\midrule
\multirow{2}{*}{Test OOD} & 250 & $4\times4$\\
 & 250 & $5\times5$\\
\bottomrule
\end{tabular}
\end{adjustbox}
\caption{Dataset statistics for STP.}
\label{tbl:datasets_stp}
\end{table}

\definecolor{codegreen}{rgb}{0,0.6,0}
\definecolor{codegray}{rgb}{0.5,0.5,0.5}
\definecolor{codepurple}{rgb}{0.58,0,0.82}
\definecolor{backcolour}{rgb}{0.95,0.95,0.92}
\lstdefinestyle{prompt}{
    language=Python,
    backgroundcolor=\color{backcolour},
    commentstyle=\color{codegreen},
    keywordstyle=\color{codepurple},
    stringstyle=\color{codegreen},
    basicstyle=\ttfamily,
    breakatwhitespace=false,
    breaklines=true,
    captionpos=b,
    keepspaces=true,
    showspaces=false,
    showstringspaces=false,
    showtabs=false,
    tabsize=2
}
\lstloadlanguages{Python}

\subsection{Prompts}
\label{sec:prompts}
The language models have been trained on a regression task with context prompts, which are provided below. Since the experiments are performed with code models, we tailor the prompt accordingly. The same prompt is used for both domains, shown in \autoref{lst:prompt}, with the puzzle representations and legend in \autoref{lst:sokoban} for Sokoban and \autoref{lst:maze}.
\subsection{Hyperparameters and Model Choice}
\label{sec:hyperparameters}
All models are trained for 40 epochs, with a learning rate of $1e-4$, batch size of 64 and optimized with Adafactor. We implement early stopping, with the model chosen by best performance on validation MAE, computed every epoch. Training is performed on 1 NVIDIA A6000 Ada GPU.

Codet5-small was chosen for experiments since, (i) it is a compute-efficient, powerful LM, and (ii) we believed the code-pretraining would be beneficial to the code-like representation of our problem.

\subsection{Additional Ablations}
\label{sec:ablations}

\paragraph{Choice of \bm{$\mathcal{C}(\cdot)$}} Theoretically, any monotonically increasing function can be used for $\mathcal{C}(\cdot)$. Practically, however, some factors need to taken care of. For instance, we cannot use $e^{g(n)}$, since it's large first derivative will assign a very high contribution value to nodes near the goal. Thus, when used for sampling, it will concentrate all the probability mass near the goal, preventing us from augmenting the training set with harder nodes, further away from the goal. 

We show additional results for two more choices for $\mathcal{C}(n)$, used in $\mathcal{D}(n, \tau)$, in \autoref{tbl:choice_c}. Note that the same $\tau$ used in the main body has been chosen, and is not tuned. Despite that, we outperform uniform sampling on most splits. This validates the general idea of using an increasing function for $\mathcal{C}(n)$. Choosing the best performing or most theoretically justified one is left for future works.

\begin{figure}[t]
\begin{center}
\begin{lstlisting}[style=prompt]
##########
#   ######
#   # ##.#
# .  $   #
#      $ #
#  #######
#@ #######
#  #######
#  #######
##########
legend =  "@ - player, # - wall, . - empty docks, ' ' - empty cell, $ - box, X - box on dock, O - player on dock"
\end{lstlisting}
\end{center}
\vspace{-1.7em}
\caption{Puzzle representation and legend of a training puzzle from Sokoban.}
\label{lst:sokoban}
\end{figure}
\begin{figure}[t]
\begin{center}
    \begin{lstlisting}[style=prompt]
#####################
#..@................#
###.#####.###.#######
#...#...#.#.#...#...#
#######.#.#.#.#####.#
#...........#.......#
###.#.#.#.#.#.#.#.#.#
#...#.#.#.#.#...#...#
#.#.#.#####.#.#.#.#.#
#.#...#.#...........#
###.#.#.#.#.###.#.#.#
#.........#...#.....#
#.#.#.#.#.#####.#.#.#
#.#.#.#.#.#.....#.#.#
#.###.#######.#.#.#.#
#...#.#X......#.....#
#.###.#.#.#.#.#.#.#.#
#...#.#.#.#.#.#.#...#
###.#####.###.#.###.#
#...#.....#...#.....#
#####################
legend =  "@ - player, # - wall, . - empty cell, X - goal"
    \end{lstlisting}
\end{center}
\vspace{-1.7em}
\caption{Puzzle representation and legend of a training puzzle from the maze dataset.}
\label{lst:maze}
\end{figure}

\begin{figure}[t]
\begin{center}
    \begin{lstlisting}[style=prompt]
puzzle_str = "i a h m v o u 0 y"
goal = "0 a h i m o u v y"
legend =  "0 - empty space"
    \end{lstlisting}
\end{center}
\vspace{-1.7em}
\caption{Puzzle representation and legend of a training puzzle from the stp dataset.}
\label{lst:stp}
\end{figure}

\begin{figure*}[t]
\begin{center}
\begin{lstlisting}[style=prompt,linewidth=\textwidth]
import torch
def get_improved_heuristic(heuristic: int, difference: int):
    '''
        A function that takes in the admissible A* heuristic and adds to it the difference, to return a heuristic closer to the optimal cost to the goal. The difference should be calculated keeping in mind the optimal cost of the puzzle.
    '''
    return heuristic + difference

# The difference is calculated by observing the {domain} puzzle and deducing the optimal cost to goal. The heuristic is subtracted from this optimal cost
# {puzzle_legend}
puzzle_str = "{puzzle_str}"
improved_heuristic = get_improved_heuristic({heuristic},
\end{lstlisting}
\end{center}
\vspace{-1.7em}
\caption{Prompt used while training the language model. \{curly braces\} denote a placeholder.}
\label{lst:prompt}
\end{figure*}

\begin{table*}[!t]
\begin{adjustbox}{width=\textwidth,center}
\begin{tabular}{c|c|cccc|cccc}
\midrule
\multicolumn{2}{c|}{\textbf{Test Splits} $\rightarrow$} & \multicolumn{4}{c|}{\textbf{IID}} & \multicolumn{4}{c}{\textbf{OOD}} \\
\midrule
\bm{$\mathcal{C}(n)$} & \textbf{Domain} & \textbf{ILR-on-solved} & \textbf{ILR-on-optimal} & \textbf{SWC} & \textbf{Optimal \%} & \textbf{ILR-on-solved} & \textbf{ILR-on-optimal} & \textbf{SWC} & \textbf{Optimal \%} \\
\midrule
$\log(\frac{|\pi|^*}{|\pi|^* - g(n)})$ & \multirow{4}{*}{Sokoban} & 10.2077 & 10.8168 & 0.9808 & 75.70 & 13.7706 & 13.7546 & 0.9828 & 77.11 \\[0.6em]
$\frac{|\pi|^*}{|\pi|^* - g(n)}$ & & 7.7467 & 7.7455 & 0.9806 & 78.87 & 11.9533 & 12.3032 & 0.9874 & 82.17 \\[0.6em]
$\frac{g(n)}{|\pi|^*}$ & & 9.2398 & 9.9242 & 0.9787 & 74.65 & 11.5371 & 11.9224 & 0.9846 & 80.24 \\
\midrule
$\log(\frac{|\pi|^*}{|\pi|^* - g(n)})$ & \multirow{4}{*}{Maze} & 1.7029 & 1.7035 & 0.9958 & 96.6 & 1.3365 & 1.3354 & 0.9964 & 95.0 \\[0.6em]
$\frac{|\pi|^*}{|\pi|^* - g(n)}$ & & 1.6119 & 1.6129 & 0.9961 & 96.6 & 1.2972 & 1.2949 & 0.9982 & 97.8 \\[0.6em]
$\frac{g(n)}{|\pi|^*}$ & & 1.6560 & 1.6553 & 0.9964 & 96.8 & 1.2691 & 1.2706 & 0.9968 & 96.2 \\
\midrule
$\log(\frac{|\pi|^*}{|\pi|^* - g(n)})$ & \multirow{4}{*}{STP}& 3.4758 & 3.9686 & 0.9765 & 73.8 & 1.4265 & 1.4606 & 0.9946 & 93.0 \\[0.6em]
$\frac{|\pi|^*}{|\pi|^* - g(n)}$ & & 3.0416 & 3.4088 & 0.9758 & 72.4 & 1.7935 & 1.8943 & 0.9885 & 86.6 \\[0.6em]
$\frac{g(n)}{|\pi|^*}$ & & 3.6157 & 4.0441 & 3.7528 & 95.4 & 1.4051 & 1.4421 & 0.9865 & 87.0\\
\bottomrule
\end{tabular}
\end{adjustbox}
\caption{Experimental results by sampling from the $\mathcal{D}(n, \tau)$, with different choices for $\mathcal{C}(\cdot)$, with the $\mathcal{L}_{L2}$ model.}
\label{tbl:choice_c}
\end{table*}

\begin{table*}[!t]
\begin{adjustbox}{width=\textwidth,center}
\begin{tabular}{c|c|cccc|cccc}
\midrule
\multicolumn{2}{c|}{\textbf{Test Splits} $\rightarrow$} & \multicolumn{4}{|c}{\textbf{IID}} & \multicolumn{4}{|c}{\textbf{OOD}}\\
\midrule
\textbf{Train Split} & \textbf{Domain} & \textbf{ILR-on-solved} & \textbf{ILR-on-optimal} & \textbf{SWC} & \textbf{Optimal \%} & \textbf{ILR-on-solved} & \textbf{ILR-on-optimal} & \textbf{SWC} & \textbf{Optimal \%}\\
\midrule
Full-data & \multirow{3}{*}{Maze} & 1.4752 & 1.4902 & 0.9925 & 94.0 & 1.2448 & 1.2467 & 0.9965 & 96.2\\
\cline{1-1} \cline{3-10}
$\mathcal{X} \sim \mathcal{U}(n)$ &  & 1.4979 & 1.5070 & \textbf{0.9897} & \textbf{92.2} & 1.1869 & 1.1769 & 0.9925 & 92.8\\
$\mathcal{X} \sim \mathcal{D}(n, 10)$ &  & \textbf{1.5517} & \textbf{1.5628} & \textbf{0.9897} & \textbf{92.2} & \textbf{1.2426} & \textbf{1.2436} & \textbf{0.9940} & \textbf{93.0}\\
\midrule
Full-data & \multirow{3}{*}{Sokoban} & 9.2978 & 10.4147 & 0.9594 & 60.92 & 14.8513 & 16.1940 & 0.9645 & 61.45\\
\cline{1-1} \cline{3-10}
$\mathcal{X} \sim \mathcal{U}(n)$ & & 7.1347 & 7.4233 & 0.9607 & \textbf{61.62} & 12.4740 & \textbf{14.7325} & 0.9500 & 48.92\\
$\mathcal{X} \sim \mathcal{D}(n, 10)$ &  & \textbf{7.8141} & \textbf{8.0857} & \textbf{0.9614} & 59.86 & \textbf{13.3144} & 12.4565 & \textbf{0.9558} & \textbf{52.53}\\
\midrule
Full-data & \multirow{3}{*}{STP} & 4.3889 & 4.9981& 0.9732 & 70.2 & 1.4297 & 1.6507 & 0.9353 & 57.0\\
\cline{1-1} \cline{3-10}
$\mathcal{X} \sim \mathcal{U}(n)$ & & 3.1497 & \textbf{3.8005} & 0.9633 & 61.2 & 1.0486 & 1.3083 & \textbf{0.9404} & \textbf{69.0} \\
$\mathcal{X} \sim \mathcal{D}(n, 3)$ & & \textbf{3.1795} & 3.7610 & \textbf{0.9662} & \textbf{63.4} & \textbf{1.0917} & \textbf{1.5482} & 0.9331 & 56.2 \\

\bottomrule
\end{tabular}
\end{adjustbox}
\caption{Experimental results with $\mathcal{L}_{LM}$ by sampling from the $\mathcal{D}(n, \tau)$ distribution. Best scores are in \textbf{bold}. }
\label{tbl:importance_sampling_lm}
\end{table*}

\subsection{Summary of Related Works} 
\label{sec:rw_summary}
A summary of the related works has been provided in \autoref{tbl:rw_summary}.

\begin{table*}[!t]
\begin{adjustbox}{width=\textwidth,center}
\begin{tabular}{p {0.23\linewidth}|p {0.4\linewidth}|p {0.39\linewidth}}
\toprule
\textbf{Research Field} & \textbf{Relevance} & \textbf{Related Works with Summary} \\
\midrule
Learning Heuristics for Planning & In this work, we make use of previous methods to learn heuristics for planning. While These primarily studied neural architectures for this problem, we fix the architecture to an LM and study the data requirements. & \textbf{Machine Learning Perspective:} These works discuss classical ML techniques to learn heuristics \cite{yoon2006learning, fern2011first, arfaee2011learning, us2013learning, chrestien2021heuristic, groshev2018learning, kirilenko2023transpath}.\\
& & \textbf{Planner Perspective:} These incorporate planner properties to learn heuristics.\cite{yonetani2021path, vlastelica2019differentiation, speck2021learning, orseau2023levin, orseau2021policy, kirilenko2023transpath,ernandes2004likely}\\
\midrule
Heuristics with LMs & The previous works studied learning heuristics with classical machine learning techniques, here we specifically discuss how LMs are used in heuristic learning. & \textbf{Tree-Search in LLMs:} These discuss how various algorithms like DFS, BFS, MCTS can be combined with LLMs for planning \cite{yao2024tree, hao2023reasoning, chen2024tree}.\\
& & \textbf{LLMs with external planners:} These discuss how symbolic solvers can be augmented with LLMs. \cite{valmeekam2023planning, gerevini2002lpg, liu2023llm+, yang2023coupling, guan2023leveraging, dagan2023dynamic}\\
& & \textbf{Improving LM-based heuristics:} These discuss how LM heuristics can be improved via training or prompting\cite{shinn2024reflexion, zhou2023language, lehnert2024beyond,gandhi2024stream}.\\
\midrule
Optimising Training Data & This is our problem statement for the planning task. & \textbf{Coreset Selection:} These works discuss the data requirements for training LMs, albeit for different tasks. To the best of our knowledge, we are the first to study coreset selection for planning \cite{paul2021deep, marion2023less, abbas2023semdedup, zhou2023algorithms,sorscher2022beyond}.\\
\bottomrule
\end{tabular}
\end{adjustbox}
\caption{A tabular summary of the related works discussed in Section \ref{sec:related-works}.}
\label{tbl:rw_summary}
\end{table*}
\end{document}